\pdfoutput=1

\documentclass[11pt]{article}

\usepackage{ACL2023}

\usepackage{times}
\usepackage{latexsym}
\usepackage{booktabs}
\usepackage{multirow}
\usepackage{tabularx}
\usepackage{graphicx}

\usepackage[T1]{fontenc}

\usepackage[utf8]{inputenc}

\usepackage{microtype}

\usepackage{inconsolata}

\title{Building blocks for complex tasks:  Robust generative event extraction for radiology reports under domain shifts}

\author{
     Sitong Zhou$^{\spadesuit}$ \quad
     Meliha Yetisgen $^{\diamondsuit}$\quad
     Mari Ostendorf$^{\spadesuit}$\quad\\
     $^\spadesuit$University of Washington, Electrical \& Computer Engineering\\
     $^\diamondsuit$University of Washington,  Biomedical and Health Informatics\\
     {\tt \{sitongz,melihay,ostendor\}@uw.edu}
}

\begin{document}
\maketitle
\begin{abstract}
This paper explores methods for extracting information from radiology reports that generalize across exam modalities to reduce requirements for annotated data.
We demonstrate that multi-pass T5-based text-to-text generative models exhibit better generalization across exam modalities compared to approaches that employ BERT-based task-specific classification layers. 
We then develop methods that reduce the inference cost of the model, making large-scale corpus processing more feasible for clinical applications. Specifically, we introduce a generative technique that decomposes complex tasks into smaller subtask blocks, which improves a single-pass model when combined with multitask training. In addition, we leverage target-domain contexts during inference to enhance domain adaptation, enabling use of smaller models. Analyses offer insights into the benefits of different cost reduction strategies.

\end{abstract}

\section{Introduction}
Radiology reports contain a diverse and rich set of clinical abnormalities documented by radiologists during their interpretation of the images. Automatic extraction of radiological findings would enable a wide range of secondary use applications to support diagnosis, triage, outcomes prediction, and clinical research \citep{Lau2020AutomaticAO}. 
We adopt an event-based schema to capture both indications, the reason for radiology exams, and abnormal findings documented in radiology reports. We use an annotated a corpus of reports from three distinct radiology examination modalities  \citep{Lybarger_rad_data}: Magnetic Resonance Imaging (MRI), Positron Emission Tomography (PET), and Computed Tomography (CT). Each event consists of a trigger, words that indicate a particular indication or finding (e.g., lesion), and a set of attributes (assertion, anatomy, characteristics, size, size trend, size count) that describe this indication or finding. Manual annotation of radiology reports is costly, therefore we hope models can generalize across different exam modalities. In this work, we define each modality in our annotated corpus as a domain and study cross-domain generalization among different modalities for the task of event extraction. Event extraction can be conceptualized as a series of subtasks, which include entity detection (trigger and attribute spans), relation detection (between triggers and attributes), and entity normalization (fine-grained labels on spans). In our experiments, we focus on trigger detection and anatomy attribute extraction with normalized labels. 

To enhance generalization capabilities, some studies employ generative models and formulate tasks as question answering and using texts to represent both inputs and outputs \citep{t5, Xie2022UnifiedSKGUA}, as opposed to allowing the model to solely learn task intent from training data \citep{spert, Lybarger_mspert_sodh}. 


The exceptional performance of generative models often rely on large model size; however, in real-time inference for processing large-scale clinical notes, reducing inference costs is crucial. To address this need, for task inference, we want to reduce the number of decoding passes and employ smaller models. 
Due to the high inference costs, there is a desire to merge these subtasks and decode them in a single step. However, the generative approach has been reported to perform better on solving subtasks individually but worsen when combined, a phenomenon referred to as the compositionality gap \citep{self_ask}. This gap can be exacerbated under domain shifts when models learn subtasks jointly, as interdependence of subtasks may vary across domains.

While large language models (LLMs)  mitigate the compositionality gap using reasoning steps \citep{chain-of-thought, self_ask} to solve complex questions by decomposing them into smaller ones, there is limited work on reasoning for highly specialized domains (such as medical event extraction) or with smaller models.  
In this paper, we reduce the compositionality gap for smaller models through formatting of complex tasks into easier subtasks as blocks. This approach teaches models how to solve individual subtasks independently and how to assemble them for solving more complex tasks. 

 The generative model enables seamless integration of supplementary contexts into the prompt, which compensates for the knowledge gap to larger models and reduces inference costs. To aid in domain adaptation, we extract target domain contexts that are likely to be helpful for the task, instead of retrieving similar contexts for general purpose. Specifically, to assist with anatomy normalization tasks, we employ an unsupervised extractor to acquire  pertinent contexts that likely contain anatomical information from the same document and/or unannotated text from the same domain. This process can either disambiguate the original single-sentence input or provide anatomy-related hints that the model can utilize. To avoid introducing source-domain-specific reliance on the contexts, we incorporate the contexts only at the inference stage.



In our experiments, we first study domain shift for extracting radiology finding events and  observe that cross-domain performance decline is more pronounced for knowledge-intensive anatomy normalization tasks, while detecting entity spans exhibits relatively stable performance. We demonstrate that building subtask blocks and assembling them as sequences to solve complex tasks can reduce the compositionality gap in smaller models. We show that incorporating target-domain contexts in domain adaptation can compensate for reduced model sizes, enabling good performance with smaller models.


\section{Task}

\begin{figure*}[htbp]
  \centering
  \includegraphics[width=\linewidth]{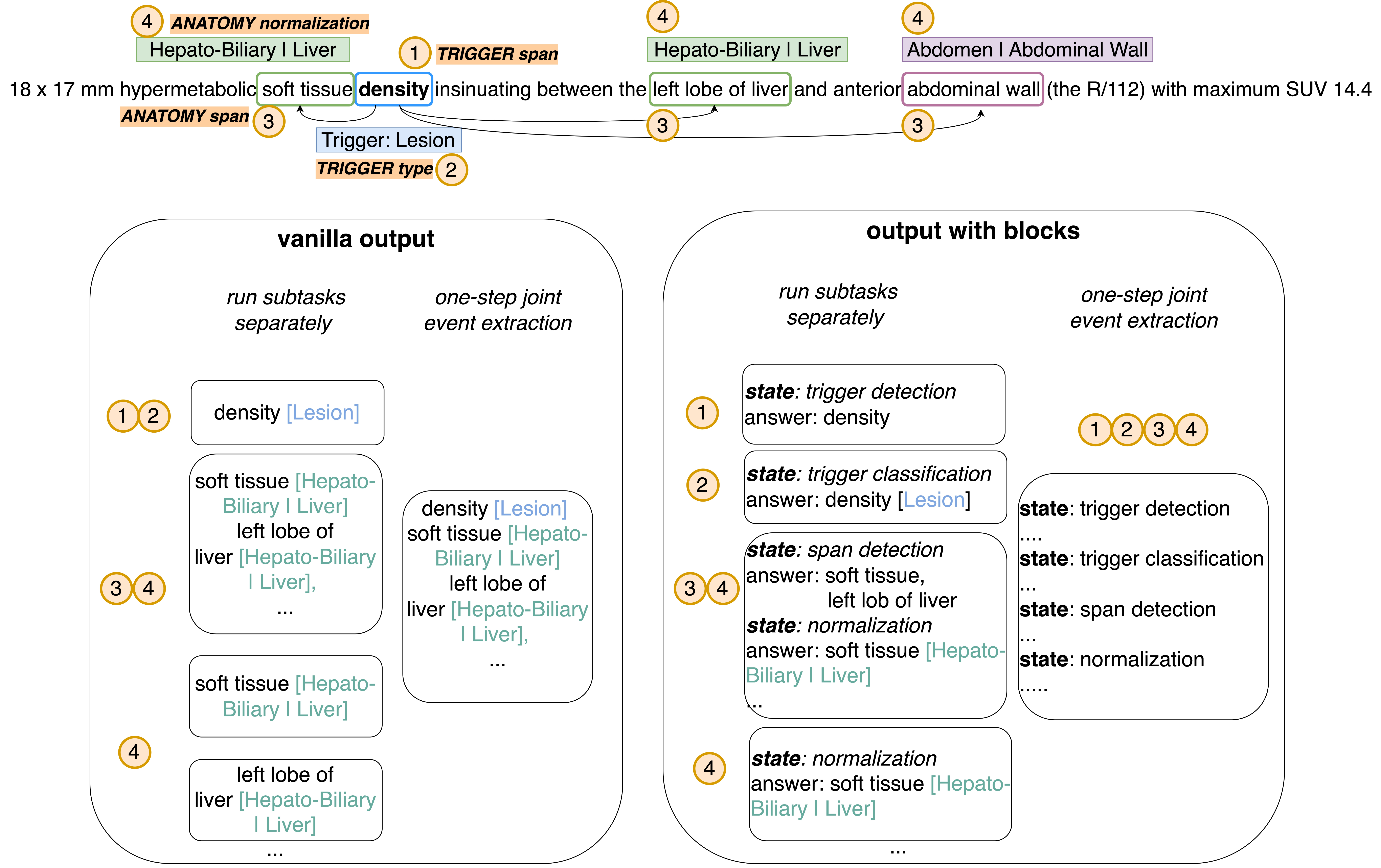}
  \caption{Representations of anatomical information in radiology reports, with the event-based annotation at the top and two generative model output formats to multi-step and one-step processing. The left-hand side shows the {\bf vanilla format} and the right-hand side shows the {\bf building block format}.} \label{fig:figure1}
\end{figure*}

\subsection{Event extraction for radiology findings }


Our event scheme includes three event types: i)~\textit{Indication} is the reason for the imaging (e.g. motor vehicle accident or cancer staging);  ii) \textit{Lesion} captures lesions uncovered by the exam (e.g. mass or tumor); and iii) \textit{Medical Problem} characterizes non-lesion abnormalities (e.g. fracture or hernia).  Each finding event is characterized by an event trigger and set of attributes (assertion, anatomy, characteristics,  
size, size-trend, count). In this work, we focus only on extracting events with normalized anatomical information and investigate cross-domain generalization for different examination modalities. Figure \ref{fig:figure1} presents a \textit{Lesion} event example. The event extraction process can be broken down into four subtasks: (1) Trigger span extraction (e.g., "density"), (2) Trigger type classification (e.g., "density" - Lesion), (3) Anatomy span extraction (e.g., "left lobe of liver" associated with the trigger "density"), and (4) Anatomy normalization to parent-child anatomy categories (e.g., "left lobe of liver" - Parent: Hepato-Biliary, Child - Liver). See Appendix~\ref{sec:appendix_anatomy_categories} for the full list of hierarchical parent-child anatomy categories.


We evaluate event extraction performance using the F1 metrics 
by \citet{radiology_kevin}. Our assessment of the trigger extraction is based on the span overlap and the event type match with respect to the gold standard labels. The anatomy extraction is first assessed at the span level. A correct anatomy prediction is associated with a correct predicted trigger and anatomy span overlap with the gold standard labels. Additionally, we evaluate anatomy extraction based on the normalization level, irrespective of their spans. A match between the predicted anatomy entity and the gold label indicates that the trigger is matched, and the normalized anatomy category is equal.

\subsection{Domain shifts across radiology modalities}




Our research investigates cross-domain generalization among three distinct radiology examination modalities: MRI, PET, and CT. These exam modalities are performed for different reasons with different technologies and the resulting radiology reports differ in terms of level of details as well as anatomy distribution.
While CT and MRI scans allow radiologists to view structures inside the body, a PET scan, on the other hand, captures how tissues in the body work on the cellular level and shows unusual activity. MRI scans very frequently involve neurological exams. The most common use of PET scans is to diagnose or monitor certain cancer types. 
In our experiments, we define each modality as a domain. We use PET as the target domain, and train on three domains separately to evaluate both in-domain and cross-domain scenarios. 


\section{Method}

\subsection{Generative event extraction with T5}
\label{sec:method_t5}
In order to improve the model's generalization capabilities over BERT-based alternatives \citep{Lybarger_mspert_sodh, spert}, we structure our event extraction task in a unified question-answering (QA) format \citep{Xie2022UnifiedSKGUA, t5}. With the generative approach, the model leverages the semantic meaning of prompts for specifying subtasks and associated categorization labels. 
Based on experiments with in-context learning \cite{Hu+2022}, we expect this to be beneficial for domain-mismatches in class label distributions, e.g. where infrequent classes in the source domain are frequent in the target domain. Furthermore, the text-to-text format offers the flexibility to incorporate additional contexts to facilitate tasks, as discussed in Section~\ref{sec:method_context}.

The input prompt comprises: (1) an input sentence from clinical notes to extract events from, (2) a question that describes the task or subtask, and (3) an ontology that provides textual labels for classification tasks and hierarchical relationships if multi-level granularities are required. 
The output is a word sequence that specifies the extracted information (the answer). Two alternative output formats are discussed in the next section; example input-output pairs for both are in Appendix~\ref{sec:appendix_t5_prompt}.

Event extraction can be seen as a multi-hop question-answering process, involving a series of subtasks for successful completion. We use a pipeline approach to address the event extraction subtasks in different steps, where each step in the pipeline consists of a specialized generative model trained for one or more of the subtask types. Three different architectures are explored: 

\textit{Three-step approach}: This involves a first step for detecting trigger spans and trigger types, followed by a second step for identifying the anatomy associated with each detected trigger, and a third step for normalizing each identified anatomical entity at parent and child levels individually. 

\textit{Two-step approach}: This involves a first step for detecting trigger spans and trigger types, followed by a second step for identifying and normalizing the anatomy associated with each detected trigger.\footnote{Both the 2-step and 3-step approaches use the same second step, predicting anatomy spans and their normalized values. The three-step approach drops the normalized values from its second step.} 

\textit{One-step approach}: we address all subtasks, which may be associated with multiple entities, in a single pass per input sentence. This method results in longer output lengths compared to the individual steps of previous two approaches.

The one-step approach substantially reduces inference costs compared to other two multi-step approaches. However, we find that it negatively impacts model performance due to the longer output and the compositionality gap. The performance loss is mostly recovered by changing the output format (as described next) together with a multi-task training strategy.  Specifically, we train the model on both the complete task and the decomposed subtasks. This allows the model to perform subtasks independently and assemble subtask sequences for complex tasks. During inference, we decode in a single step to minimize costs. 


Our work builds on generative models, specifically the clinical version of the T5 models \citep{clinicalt5}, which are pre-trained on medical articles and clinical notes. This choice leverages their strengths in comprehending clinical text styles and medical knowledge.

\subsection{Output formats}
\label{sec:method_outputs}

We explore two different output formats as illustrated in Figure~\ref{fig:figure1}, with subtask answers provided in sequence when there are multiple subtasks.

The baselines leverage a standard output format (referred to here as the \textbf{vanilla format}), which specifies the answer for a subtask with an extracted span followed by the entity label in brackets "[]". When multiple entities are detected, they are generated in sequence.

The vanilla format can be used with the one-step approach, but the resulting output can be very long when multiple triggers and/or entities are detected. The lack of distinction between types of spans in the output makes it harder for the language model to learn the subtask structure.  To address this problem, we introduce a state-augmented prompt (referred to as the \textbf{building block} format), in which each subtask is associated with a state (as in a finite-state transducer) and explicitly named.
Our approach is motivated by the work on chain-of-thought LLMs \citep{chain-of-thought, self_ask}, which use natural language reasoning in the generated outputs to address the compositionality gap. However, it differs in that we do not use natural language reasoning, but rather more of a programming-like description. In addition, the finite-state framework is amenable to meulti-task training, which is particularly important for the block approach.



\subsection{Using target-domain contexts in prompts}
\label{sec:method_context}

A single input sentence may not provide enough information for a model to complete a task, as additional details may be needed for disambiguation or to supplement missing knowledge in pre-trained language models. 
Fortunately, the text format of the input allows for the seamless integration of additional contexts from the target domain during inference to aid in the task and infuse helpful domain-specific bias, even if the models were not trained for reading contexts. 

The desired contexts should be relevant to the input sentence and contain helpful task information. We utilize two types of contexts: document-level and domain-level contexts to help anatomy normalization subtasks. Document-level contexts include adjacent sentences before and after the input, automatically extracted section headers\footnote{We extract section headers as the beginning of the last previous sentence containing ':'} and exam type metadata associated with the same clinical note.  The document-level contexts are likely to describe relevant anatomical parts, as section headers and exam types often summarize anatomical information. Domain-level contexts are retrieved from the unlabeled target-domain corpus.  We search for the most similar sentence with the greatest lexical overlap degree, using the search algorithm BM25 \citep{bm25_python}.\footnote{We implement the BM25 algorithm using \url{https://github.com/dorianbrown/rank_bm25}} When the search pool is large, the top-ranked retrieved context sentence likely describes a similar anatomy part as the queried input sentence. To reduce computational costs and ensure that the retrieved sentences contain useful anatomical information, we pre-filter the target corpus to limit the search scope to sentences containing common anatomy terms listed from anatomy normalization categories and high-frequency auto-extracted section headers, reducing the number by 74\%.  More context-retrieval details are in Appendix~\ref{sec:appendix_context_retrieval}.

We add contexts only during decoding (and not in training) to prevent the model from relying too much on source-domain contexts. In the input prompts, exam type, section headers and prior sentences are placed before input sentences, following their natural orders. Other contexts are inserted between the input sentences and task ontology.\footnote{The full T5 input template is described in Table~\ref{tab:prompt_examples} from Appendix~\ref{sec:appendix_t5_prompt}} We test this approach in a separate anatomy normalization run after the one-step building block model. This process combines building block output format with target domain context integration. The reason for not directly adding it to a one-step process is that introducing contexts to inputs can potentially corrupt span detection, as the model may extract spans from the context rather than exclusively from the input sentence.

\section{Experiments}
\subsection{Radiology datasets across exam modalities}


\begin{table}[h!]
\centering
\begin{tabular}{@{}rcc@{}}
\toprule
\textbf{Data split} & \multicolumn{1}{l}{\textbf{Note Count}} & \multicolumn{1}{l}{\textbf{Sent Count}} \\ \midrule
CT (train)        & 143                                   & 3707                                 \\
MRI (train)        & 144                                   & 3551                                 \\
PET (train)        & 142                                   & 5184                                 \\
PET (valid)          & 20                                    & 758                                  \\
PET (test)         & 40                                    & 1481                                 \\
PET (unlabeled)         & 1471                                 & 50000                                 \\\bottomrule
\end{tabular}
\caption{Dataset statistics for the three radiology examination modalities: CT, MRI, and PET. We explore in-domain and cross-domain training, evaluating on PET. }
\label{tab:dataset_statistics}
\end{table}

We use 
an annotated corpus containing radiology notes 
about CT, MRI, and  PET imaging exams; statistics are given in Table~\ref{tab:dataset_statistics}. 
The anatomy normalization labels are grouped into sublevels according to the SNOMED CT concepts. Notes in the test and validation sets are all doubly annotated.  The inter-rater agreement for Trigger is 0.73 F1. 

Variations in anatomy distribution across imaging modalities can cause domain discrepancies. PET has the most balanced distribution among parent-level anatomy categories, followed by CT. However, MRI has a heavily skewed distribution, with 62\% of trigger-associated anatomy entities being neurological among 16 parent-level categories.
See Appendix~\ref{sec:appendix_anatomy_categories} for anatomy distribution details.

To enhance domain-specific context retrieval and boost the chances of retrieving helpful contexts, we expand the search pool by sampling 50,000 unlabeled PET report sentences from the same distribution as in the annotated reports \citep{Lybarger_rad_data}, with a minimum of three tokens.


\subsection{Implementation}


In the non-generative baseline, we adopt the mSpERT model \citep{Lybarger_mspert_sodh} for hierarchical multi-label entity and relation extraction. Entities are extracted as spans. We initialize with Bio-Clinical BERT \citep{bioclinicalbert}. 

For the T5 model using both vanilla output formats and the subtask block formats, we initialize with ClinicalT5 \citep{clinicalt5}, 

For all models, the best checkpoint is chosen after 15 training epochs based on the validation performance on the target domain. For T5 models with multitask training on subtask blocks, which involves a higher number of training steps, we evaluate the model on the validation set after every 0.5 epoch approximately. For methods that do not involve multitask training, we evaluate the model on the validation set per epoch.

We implement multitask training on subtask blocks for MRI and PET, using the auxiliary tasks, as described in Section~\ref{sec:method_t5}, including trigger span detection, trigger classification, joint anatomy span detection and normalization, and anatomy normalization.  For the CT-PET transfer scenario, we add an additional anatomy span detection auxiliary task, as we observe that more aggressive learning is needed for anatomy span detection in the CT domain. Detailed information about hyperparameters can be found in Appendix~\ref{sec:appendix_implementation}.
\begin{table*}[ht]
\centering
\caption{F1 scores (\%) for: non-generative mSpERT \cite{Lybarger_mspert_sodh}, generative vanilla T5 models with both multi-step pipeline and one-step joint approaches, and our proposed one-step T5 model using the building block technique. All models adopt the T5-base architecture and are initialized with ClinicalT5 \citep{clinicalt5}.  \textbf{Best overall} scores are in bold, and \underline{best one-step} scores are underlined.}
\label{tab:main}
\begin{tabularx}{\textwidth}{@{}l*{5}{>{\centering\arraybackslash}X}@{}}
\toprule

\textbf{Entity} & \textbf{mSpERT} & \textbf{T5-base 3-step (vanilla)} & \textbf{T5-base 2-step (vanilla)} & \textbf{T5-base 1-step (vanilla)} & \textbf{T5-base 1-step (blocks)} \\ \midrule
& \multicolumn{5}{c}{\textbf{PET → PET}} \\ \cmidrule(lr){2-6}
Trigger & 82.4 & 81.9 & 81.9 & 82.1 & \textbf{82.6} \\
Anatomy Span & 65.8 & \textbf{67.6} & \textbf{67.6} & 66.0 & \underline{66.1} \\
Anatomy Parent & 61.9 & 64.7 & \textbf{64.9} & 63.3 & \underline{63.5} \\
Anatomy Child & 59.6 & 62.1 & \textbf{62.3} & 59.7 & \underline{60.7} \\
\midrule
& \multicolumn{5}{c}{\textbf{MRI → PET}} \\ \cmidrule(lr){2-6}

Trigger & 75.6 & 76.6 & 76.6 & 76.4 & \underline{\textbf{77.8}} \\
Anatomy Span & 59.9 & 60.9 & 60.9 & 59.2 & \underline{\textbf{61.1}} \\
Anatomy Parent & 34.7 & \textbf{48.6} & 47.1 & 44.9 & \underline{48.3} \\
Anatomy Child & 33.5 & 44.6 & 44.0 & 41.2 & \underline{\textbf{44.8}} \\
\midrule
& \multicolumn{5}{c}{\textbf{CT → PET}} \\ \cmidrule(lr){2-6}

Trigger & 75.7 & 76.1 & 76.1 & 74.0 & \underline{\textbf{76.6}} \\
Anatomy Span & 59.7 & \textbf{61.4} & \textbf{61.4} & 56.3 & \underline{59.8} \\
Anatomy Parent & 53.2 & \textbf{55.8} & 54.8 & 50.8 & \underline{55.0} \\
Anatomy Child & 47.5 & \textbf{53.3} & 51.8 & 48.1 & \underline{51.2} \\
\bottomrule
\end{tabularx}
\end{table*}

\section{Results}


Table~\ref{tab:main} shows the trigger and anatomy detection results for mSpERT compared to different context-independent T5-base alternatives. For the in-domain condition, all T5 approaches outperform the mSpERT model for the three anatomy-related metrics. The results for trigger detection are mixed, but fairly similar for all. The best performance overall is obtained using the 2-step vanilla output T5 model. 
For the cross-domain scenarios, all models suffer degradation in performance compared to the in-domain condition, with the greatest performance drop for the normalized anatomy categories, particularly for the MRI-PET condition which has the greatest mismatch in anatomy distribution.
The performance loss is greatest for the mSpERT model, with a 44\% relative reduction in F1 scores for normalized anatomy (at both parent and child levels) for the MRI-PET case. 
In contrast, the relative loss on the parent and child levels for the T5 models is 24-29\%. 
For both within and across-domain scenarios, the building block technique improves the 1-step results for all categories, but particularly for the more difficult anatomy normalization tasks.
As described later in Section~\ref{sec:analyze_compositionality}, the 1-step approach is sensitive to the compositionality gap, which is ameliorated by the block approach.
For the cross-domain scenarios, the best overall results are obtained with the 3-step approach for the CT-PET condition and with the 1-step block approach for the MRI-PET condition (greater mismatch). An additional advantage of the 1-step approach is the lower latency associated with using only one decoding pass.




\begin{table*}[ht]
\centering
\caption{F1 scores (\%) for T5 anatomy classification models with and without contexts. Results with context involve a first pass with the 1-step T5-base building blocks method,  the same as "T5-base
one-step
(blocks)" in Table~\ref{tab:main}, followed by another pass that normalizes the anatomy spans that are previously detected by the 1-step T5-base (block) model. We normalize with the model used in the last step of the 3-step (vanilla) pipeline, optionally augmented with contexts in the prompts. We also add the T5-large normalization model without context to compare with the larger-scale counterpart. }
\label{tab:contexts}
\begin{tabularx}{\textwidth}{@{}l*{6}{>{\centering\arraybackslash}X}@{}}
\toprule
\textbf{Normalization model} & \textbf{T5-large} & \textbf{T5-base} & \textbf{T5-base } & \textbf{T5-base } & \textbf{T5-base} & \textbf{T5-base } \\ 
\midrule
\textbf{Context} & \textbf{n/a} & \textbf{n/a} & \textbf{ adjacent sentences} & \textbf{  metadata \&  header } & \textbf{  BM25 retrieval} & \textbf{  all combined} \\ \midrule

& \multicolumn{6}{c}{\textbf{PET → PET},    Trigger: 82.6, Anatomy Span: 66.1} \\ \cmidrule(lr){2-7}
Anatomy Parent & 63.6 & \textbf{63.9} & 63.8 & 63.7 & 63.8 & 63.7 \\
Anatomy Child & 60.9 & 60.9 & 61.0 & \textbf{61.1} & 60.3 & 60.4 \\
 \midrule
& \multicolumn{6}{c}{\textbf{MRI → PET},     Trigger: 77.8, Anatomy Span: 61.1} \\ \cmidrule(lr){2-7}
Anatomy Parent & 51.2 & 50.8 & 52.1 & 51.6 & \textbf{53.8} & 53.5 \\
Anatomy Child & 48.6 & 45.4 & 47.1 & 46.6 & 48.3 & \textbf{48.8} \\
 \midrule
& \multicolumn{6}{c}{\textbf{CT → PET}, Trigger: 77.8, Anatomy Span:   59.8 } \\ \cmidrule(lr){2-7}
Anatomy Parent & 54.1 & 54.2 & 55.5 & 55.0 & 55.5 & \textbf{55.9} \\
Anatomy Child & 51.2 & 51.2 & 52.2 & 51.6 & 52.6 & \textbf{53.0} \\
\bottomrule
\end{tabularx}
\end{table*}

As described earlier, target-domain contexts are added to prompts during a second step of T5 decoding to help anatomy normalization, after the 1-step subtask block decoding with T5-base. Table~\ref{tab:contexts} shows results for all different types of contexts, as well as using either T5-large or T5-base in the second step without context.
Without context, the T5-base and T5-large models give similar results for in-domain and CT-PET cross-domain conditions, but T5-large improves results for the MRI-PET condition. (Note that T5-large is only used in the last step; a bigger benefit could be observed if used in both steps.) All types of context are useful for the two domain-shift cases, but there is little or no benefit for the in-domain case.
Of the different types of context, automatically retrieved similar sentences from unlabeled target-domain data provide the greatest benefit in the mismatched scenarios. Combining all contexts provides a small additional benefit, except for the anatomy parent in the MRI-PET case.
Anecdotally, we observe that same-document contexts are useful for disambiguation, while hints for challenging examples are more likely collected from a large domain-level corpus rather than just the same document. (For examples, see Appendix~\ref{sec:appendix_case_study}.)\\\\
Table~\ref{tab:cost_inference2} provides information on the relative cost of the different T5 models. The multi-pass models have higher latency (average passes/sample) in that passes are necessarily sequential. (Note that samples with no findings or no anatomy identified in the first pass do not require additional passes.) The number of tokens per sample is an indicator of cost. The 1-step model with blocks has a higher cost than the 2-step approach because of the additional tokens introduced by the state-augmented prompt, but the cost is still lower than the 3-step approach. The use of context adds additional cost. 

\begin{table}[ht]
\centering
\caption{Average number of decoding passes per 
sample (indicating relative decoding time) and tokens per sample (indicating relative cost) of one-step and multi-step approaches for testing on the PET domain. The token counts per sample are the average of the sum of input and output token counts, which is used for proportionality pricing LLM usage by ChatGPT. The context method uses all context combined in another normalization step as in Table~\ref{tab:contexts}. }
\label{tab:cost_inference2}
\begin{tabular}{lcc}
\toprule
 & \textbf{passes/} & \textbf{tokens/} \\ 
 \textbf{Method} & \textbf{sample} & \textbf{sample} \\ \midrule
3-step (vanilla) & 2.5 & 355 \\
2-step (vanilla) & 1.7 & 199 \\
1-step (block) +  context& 1.7 & 450 \\
1-step (block) & 1 & 245 \\
\bottomrule
\end{tabular}
\end{table}

\section{Analysis}


In this section, we analyze results to better understand performance improvements associated with the subtask block format and retrieved context in prompts.

\subsection{Multitask training for subtask blocks}
\label{sec:ablation_multitask}

\begin{table}[ht]
\centering
\caption{F1 scores (\%) for the cross-domain MRI-PET condition using 1-step T5-base models, comparing: vanilla output format, building block format but no multitask training, and building block format with multitask training.}
\label{tab:ablation_multitask}
\begin{tabularx}{\columnwidth}{@{}>{\hsize=.8\hsize}X*{3}{>{\centering\arraybackslash\hsize=1.06666\hsize}X}@{}}
\toprule
\textbf{Entity} & \textbf{vanilla} & \textbf{blocks, no multitask} & \textbf{blocks, multitask} \\ \midrule
Trigger & 76.4 & 76.0 & \textbf{77.8} \\
Anatomy & 59.2 & 57.1 & \textbf{61.1} \\
Parent & 44.9 & 38.6 & \textbf{48.3} \\
Child & 41.2 & 36.9 & \textbf{44.8} \\
\bottomrule
\end{tabularx}
\end{table}

To understand the contributing factors for the subtask block method's effectiveness, we examine whether the output format encodes helpful structural task information, or multitask training on individual subtasks predominantly drives performance. We conduct an additional experiment using the same subtask block output format, but without the multitask training for individual blocks. We use MRI as the source domain, because it suffers the most cross-domain performance drop.  The results in Table~\ref{tab:ablation_multitask} show a substantial drop in the model's performance in the absence of multi-task training, as compared to both the multi-task version and the baseline output format. This performance degradation may be attributed to increased decoding lengths.

\subsection{Predictions for multiple anatomy parents}
\label{sec:analyze_compositionality}

In addition to differences in the anatomy parent class distribution across domains, the three examination modalities also differ in how frequently sentences with multiple anatomy entities involve multiple parent classes.  As shown in Table~\ref{tab:analyze_compositionality}, 57\% of the sentences with multiple anatomy entities in the target domain (PET) have multiple parents, whereas the percentage is much lower for the other domains (only 12\% for MRI). When using the vanilla method, models trained on a domain with few instances of multiple parents will tend to predict the same parent class for each entity, as shown by the lower frequency of prediction in the table. The use of subtask blocks together with multitask training substantially improves the model's ability to identify multiple parent types when there are multiple anatomy entities.  In all domains, roughly 20\% of sentences have multiple anatomy entities, so this leads to overall performance improvement.


\begin{table}[ht]
\centering
\caption{Relative frequency (\%) of sentences with multiple anatomy entities that have different parents, comparing frequencies as predicted by different models to the frequencies based on gold annotations for training data. The gold relative frequency on the PET test data is 55\%.}
\label{tab:analyze_compositionality}
\begin{tabularx}{\columnwidth}{@{}l*{3}{>{\centering\arraybackslash}X}@{}}
\toprule
\textbf{Domain} & \textbf{Training} & \textbf{Vanilla} & \textbf{Blocks} \\
\midrule
PET & 57 & 53 & \textbf{56} \\
MRI & 12 & 29 & \textbf{46} \\
CT & 33 & 45 & \textbf{52} \\
\bottomrule
\end{tabularx}
\end{table}


\subsection{Target domain retrieval filtering}
\begin{table}[ht]
\centering
\caption{Normalized anatomy F1 score (\%) for the MRI-PET condition, comparing approaches for using target-domain context retrieved using BM25: no context, unfiltered retrieval, and filtering the retrieval corpus to anatomy informative sentences.}
\label{tab:ablation_contexts}
\begin{tabularx}{\columnwidth}{@{}>{\hsize=.8\hsize}X*{3}{>{\centering\arraybackslash\hsize=1.06666\hsize}X}@{}}
\toprule
\textbf{Entity} & \textbf{no context} & \textbf{unfiltered contexts} & \textbf{filtered contexts} \\ \midrule
Parent & 50.8 & 52.7 & \textbf{53.8} \\
Child & 45.4 & 47.4 & \textbf{48.3} \\
\hline
 \multicolumn{4}{c}{Trigger: 77.8, Anatomy: 61.1 } \\
\bottomrule
\end{tabularx}
\end{table}

To reduce the search costs, we filter the unlabeled target domain data to include only sentences with anatomy terms before running retrieval with BM25. To understand the impacts on performance, we run experiments on unfiltered data, again focusing on the MRI data where domain differences are greatest. Table~\ref{tab:ablation_contexts} shows that filtering for anatomy not only reduces costs but also gives a small improvement in results for identifying normalized categories.



\section{Related work}
\subsection{Event extraction methods}
Event extraction research has predominantly depended on BERT-based \citep{bert, bioclinicalbert} models, where the extraction subtasks are performed by classifiers utilizing the language model layer representations \citep{ spert, pure, Lybarger_mspert_sodh}. They often yield satisfactory results when training on sufficient in-domain training data.
 For example, when training and testing on CT scan reports, normalizing anatomical terms can result in an F1 score of 79\% for nine major body parts and 73\% for 41 sub-body parts \citep{radiology_kevin}. 
Recently, there has been growing interest in adopting generative approaches \citep{t5, gpt3} for information extraction, which incorporates task descriptions and auxiliary context information to enhance performance \citep{Xie2022UnifiedSKGUA}. Many efforts \citep{clinicalt5,scifivet5,clinicalt5_mimic4,biogpt} support exploration of clinical tasks through pre-training generative models for biomedical and clinical domains.  In this study, we explicitly evaluate generative models in domain shift settings,  with an emphasis on minimizing inference costs.

\subsection{Context augmentation}

Integrating models with supplementary contexts has shown benefits in knowledge-intensive tasks \citep{rag, realm}. Generative models can utilize knowledge prompts from external knowledge sources \citep{baolin_fact_check_external_knowledge,gary_gpt3_knowledge}. In our work, we retrieve contexts from the unlabeled clinical note corpus without relying on external resources.


\subsection{Compositionality Gap}
The compositionality gap has been identified as a challenge in generative models when multiple subtasks are combined \citep{self_ask}. Prior research on large language models has demonstrated that breaking down complex tasks into smaller subproblems can be beneficial \citep{chain-of-thought, self_ask}. 
Small models have been employed for multiple decoding passes \citep{text_modular_network}, but there is limited research on reasoning with smaller models that merge these steps, which is essential for real-time applications in the clinical field.
\section{Conclusion}
In conclusion, we present generative event extraction methods for radiology findings that improve generalization under domain shifts and reduce the inference costs. By decomposing complex tasks into simpler subtask blocks and incorporating target-domain context during the inference process, our approach enables smaller models to achieve performance similar to or better than those obtained with more decoding passes, and comparable to larger models on anatomy normalization. Our methods make efficient inference for extensive clinical notes more feasible. This work offers insights into reasoning with smaller models and using context to compensate the reduced model size.


\section*{Limitations}
The use of machine learning models in clinical decision-making requires an understanding of the reasoning behind model predictions. Our study focuses on improving the performance of smaller models using context and subtask blocks. While the subtask state labels provide some interpretability, we have not explored its impact on trust among medical professionals. 
In addition, the relative benefit of the different multi-pass strategies and different types of context appear to depend on the degree of domain mismatch, which should be further explored in future work.

\section*{Ethics Statement}
 Radiology reports contain sensitive patient information and it is crucial to handle this data responsibly, adhering to strict privacy and confidentiality guidelines. The dataset used in this paper was fully-de-identified. We received approval from our institution's IRB prior to conduct the presented research and used HIPAA compliant servers. Additionally, a careful examination is needed to assess potential bias in models used for extracting information from radiology reports prior to implementing real life secondary use applications.

\section*{Acknowledgements}
This work was supported by NIH/NCI (1R01CA248422-01A1 and 1R21CA258242-01).

\bibliography{anthology,custom}
\bibliographystyle{acl_natbib}

\appendix
\section{Hierarchical anatomy normalization categories}
\label{sec:appendix_anatomy_categories}

\begin{table*}[h!]
\centering
\begin{tabular}{p{3cm}p{10cm}}
\toprule
Parent-level Class & Child-level Classes \\
\midrule
Neurological & Undetermined, Spine Cervical, Spine Thoracic, Spine Lumbar, Spine Sacral, \\
             & Spine Cord, Spine Unspecified, Brain, Nerve, Pituitary, Cerebrospinal Fluid Pathway, \\
             & Cerebrovascular System, Extraaxial \\ \addlinespace
Cardiovascular & Undetermined, Venous, Arterial, Pulmonary Artery, Heart, Pericardial Sac, \\
               & Coronary Artery \\ \addlinespace
Thoracic & Undetermined, Mediastinal \\ \addlinespace
Respiratory & Undetermined, Lung, Pleural Membrane, Tracheobronchial \\ \addlinespace
Digestive & Undetermined, Esophagus, Stomach, Intestine, Small Intestine, \\
          & Large Intestine \\ \addlinespace
Hepato-Biliary & Undetermined, Gallblader, Bile Duct, Pancreas, Liver \\ \addlinespace
Urinary & Undetermined, Kidney, Urinary Bladder, Ureter \\ \addlinespace
Lymphatic & Undetermined \\ \addlinespace
F Reproductive Obstetric & Undetermined, Breast, Ovary, Uterus, Adnexal, Extra-embryonic, \\
                         & Placenta, Fetus, Umbilical Cord, Female Genital Structure \\ \addlinespace
M Reproductive & Undetermined, Prostate, Testis, Epididymis \\ \addlinespace
Musculo-Skeletal & Undetermined, Skeletal and or Smooth Muscle, Bone and or Joint \\ \addlinespace
Body Regions & Undetermined, Entire Body, Pelvis, Lower Limb, Upper Limb \\ \addlinespace
Head Neck & Undetermined, Thyroid, Neck, Ear, Eye, Mouth, Nasal Sinus, \\
          & Pharynx, Laryngeal \\ \addlinespace
Skin & Undetermined, Skin and or Mucous Membrane, Subcutaneous \\ \addlinespace
Abdomen & Undetermined, Retroperitoneal, Abdominal Wall, Peritoneal Sac, \\
        & Spleen, Adrenal Gland, Mesentery \\ \addlinespace
Miscellaneous & Undetermined, Adipose Tissue, Connective Tissue, Biomedical Device \\
\bottomrule
\end{tabular}
\caption{Hierarchical anatomy normalization categories at parent and child levels.}
\label{tab:anatomy_ontology}
\end{table*}

\begin{figure*}[htbp]
  \centering
  \includegraphics[width=0.8\linewidth]{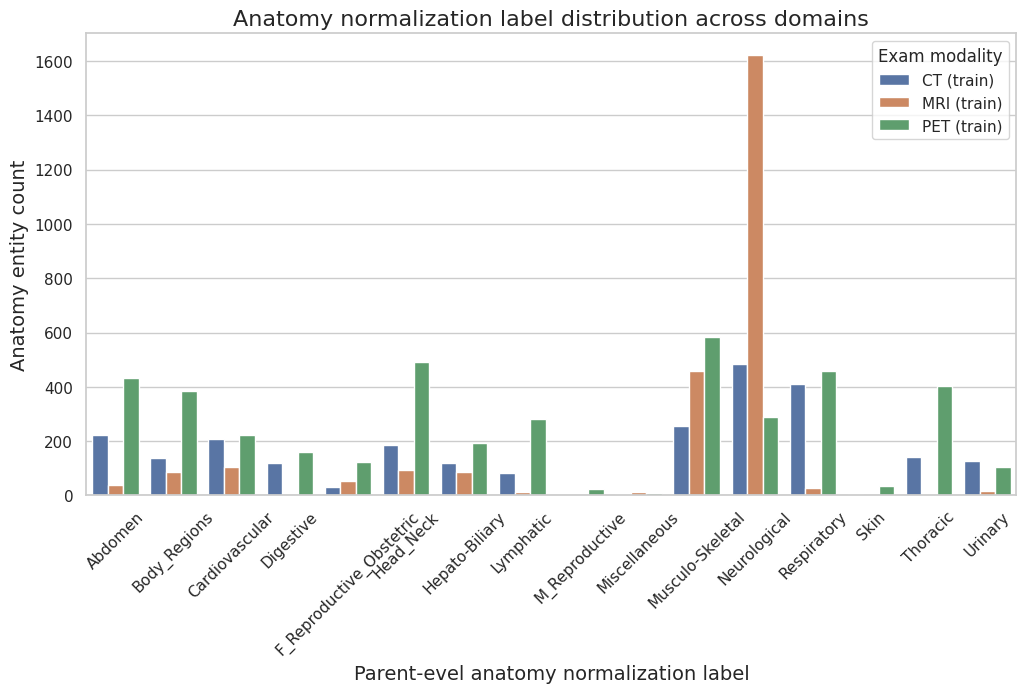}
  \caption{Domain differences in terms of the frequencies of parent-level anatomy normalization labels from the training data. }
  \label{fig:cross_domain_anatomy_dist}
\end{figure*}

We normalize detected anatomy spans for applications focusing on anatomy categories rather than specific anatomy terms. We classify at different granularities, a parent-level coarse classification with 16 parent labels and a child-level fine-grained classification with 72 categories. Each parent-level class includes an "Undetermined" child-level class to account for cases that don't fit into its other specified child classes. The full normalization categories are in Tabel~\ref{tab:anatomy_ontology}.

As shown in Figure~\ref{fig:cross_domain_anatomy_dist},
MRI exhibits a more imbalanced distribution, with a majority of the anatomies related to the "Neurological" parent-level class.  
 In CT exams, "Respiratory" account for 16\% and "Neurological"  represent 19\% among all finding-related anatomies. For MRI, "Musculo-Skeletal"  constitutes 18\% while "Neurological" exams make up a substantial 62\%. Lastly, in PET, "Head Neck" accounts for 12\% and "Musculo-Skeletal"  comprises 14\%.
 

\section{Generative method input and output formats}
\label{sec:appendix_t5_prompt}
\begin{table*}
    \centering
\begin{tabular}{p{0.15\textwidth}|p{0.35\textwidth}|p{0.4\textwidth}}
     \hline
   \multicolumn{3}{p{15cm}}{\textbf{Template}: [Input sentence] [Question] structured knowledge: [Task ontology]	}\\
     \multicolumn{3}{p{15cm}}{\textbf{Template with contexts}: [Prepended contexts] [Input sentence] [Question] structured knowledge: [Appended contexts] [Task ontology]	}\\
   \hline
\multicolumn{3}{p{15cm}}{\textbf{Trigger task ontology} (for T5-vanilla pipeline: 1st step)	Indication | Lesion | Medical\_Problem}\\
	\multicolumn{3}{p{15cm}}{\textbf{Anatomy task ontology}  (for T5-vanilla pipeline: 2nd step, 3rd step) Neurological: Undetermined, Spine Cervical, Spine Thoracic .... (see Table~\ref{sec:appendix_anatomy_categories})}\\
\multicolumn{3}{p{15cm}}{\textbf{Trigger anatomy task ontology} (for all related to one-step building block ) 	 trigger types: Indication | Lesion | Medical\_Problem 
 anatomy categories: Neurological: Undetermined, Spine, ...} \\
   \hline
 \multicolumn{3}{p{15cm}}{\textbf{Input sentence}: 18 x 17 mm hypermetabolic soft tissue density insinuating between the left lobe of the liver and anterior abdominal wall ( the R/112 ) with maximum SUV 14.4 .	} \\
        \hline
        \textbf{Model} & \textbf{Question} & \textbf{Output} \\
        \hline
        T5-vanilla pipeline: first step (trigger span \& type) & Question: What are medical findings in this sentence? & trigger: density [ Lesion ] \\
        \hline
second step  (anatomy span \& type) & Consider the medical finding "density" in the span "hypermetabolic soft tissue density insinuating between the", Question: What anatomy it occurs in?  Where is it located? & anatomies: soft tissue [ Hepato-Biliary | Liver ], left lobe of the liver [ Hepato-Biliary | Liver ], anterior abdominal wall [ Abdomen | Abdominal Wall ] \\
        \hline
third step (anatomy normalization) & Consider the anatomy "soft tissue" in the span "17 mm hypermetabolic soft tissue density insinuating between", which anatomy category it belongs to among listed options? & anatomies: soft tissue [ Hepato-Biliary | Liver ] \\
        \hline
                T5-vanilla one-step (trigger span \& type, anatomy span \& normalization) & Question: What are medical findings in this sentence? What anatomy they occur in?  which anatomy category they belong to among listed options? & trigger: density [ Lesion ] anatomies: soft tissue [ Hepato-Biliary | Liver ], left lobe of the liver [ Hepato-Biliary | Liver ], anterior abdominal wall [ Abdomen | Abdominal Wall ] \\
        \hline
        T5 one-step subtask blocks (trigger span \& type, anatomy span \& normalization) & [ same as T5-vanilla one-step ] & state: trigger detection answer: density state: trigger classification answer: density [ Lesion ] state: span detection answer: soft tissue, left lobe of the liver, anterior abdominal wall state: classification answer: soft tissue [ Hepato-Biliary | Liver ] state: classification answer: left lobe of the liver [ Hepato-Biliary | Liver ] state: classification answer: anterior abdominal wall [ Abdomen | Abdominal Wall ] \\
        \hline
        multitask for trigger classification  (trigger type) & Consider the medical finding "density", Question: What is the type of this medical finding? & state: trigger classification answer: density [ Lesion ] \\
        \hline
         multitask for anatomy span (anatomy span) & Consider the medical finding "density" in the span "hypermetabolic soft tissue density insinuating between the", Question: Please identify terms that describe the finding's anatomy locations. & state: span detection answer: soft tissue, left lobe of the liver, anterior abdominal wall \\
        \hline
    \end{tabular}
    \caption{Templates and examples for T5 inputs and outputs.  The "multitask" rows correspond to auxiliary tasks for the T5 one-step subtask block method. We omit rows for "multitask for anatomy" and "multitask for anatomy normalization", since they use the same question format as the 2nd and 3rd steps of the pipeline approach, but with answers in the subtask block format.}

    \label{tab:prompt_examples}
\end{table*}
We document the templates for the input and output, with examples in Tabel~\ref{tab:prompt_examples}. For the template with contexts, "prepended contexts" include prior sentences, section headers, exam type metadata, other contexts are "appended contexts". 
\section{Post-processing for the generative event extraction}
\label{sec:appendix_post_processing}

When matching spans in the input sentence for predicted terms, for single-token terms, we match the corresponding token. For multiple-token phrases, we match phrases using the longest common normalized string to the input sentence. Where multiple matches are found, we choose the first match from the left of the sentence, while for anatomy spans, we choose the closest match to their query triggers.
\section{Domain-level context retrieval}
\label{sec:appendix_context_retrieval}

We conduct a domain-level context search using 50,000 sentences from the target domain (PET) corpus with more than three tokens, plus 1841 sentences from the test set. The retrieved content must not be the input sentence itself. For each input clinical sentence, we identify the most lexically similar sentence from the search pool by selecting the one with the highest BM25 score. We remove punctuation and lowercase each input query when matching it with the search corpus sentences using the BM25 method.

\begin{table*}[ht]
\centering
\begin{tabular}{p{15cm}}
\toprule
Neurological: Spine Cervical, Spine Thoracic, Spine Lumbar, Spine Sacral, Spine Cord, Spine, Brain, Nerve, Pituitary, Cerebrospinal, Cerebrovascular, Extraaxial \\
Cardiovascular: Venous, Arterial, Pulmonary Artery, Heart, Pericardial Sac, Coronary Artery\\
Thoracic: Mediastinal\\
Respiratory: Lung, Pleural Membrane, Tracheobronchial\\
Digestive: Esophagus, Stomach, Intestine, Intestine, Intestine\\
Hepato-Biliary: Gallbladder, Bile, Pancreas, Liver\\
Urinary: Kidney, Urinary Bladder, Ureter\\
Reproductive: Breast, Ovary, Uterus, Adnexal, Extra-embryonic, Placenta, Fetus, Umbilical Cord, Genital Structure, Prostate, Testis, Epididymis\\
Musculo-Skeletal: Skeletal, Smooth Muscle, Bone, Pelvis, Limb\\
Head Neck: Thyroid, Neck, Ear, Eye, Mouth, Nasal Sinus, Pharynx, Laryngeal\\
Skin: Skin, Mucous Membrane, Subcutaneous\\
Abdomen: Retroperitoneal, Abdominal, Peritoneal Sac, Spleen, Adrenal, Mesentery,\\ Adipose, Chest, Mediastinum, Osseous, Bones, Extremities, Lungs, Musculoskeletal, Ventricular, Bowel, Pleura, Spleen, Vasculature, Thorax, Gallbladder, Kidneys, Adrenals, Adrenal, Cardio \\
\bottomrule
\end{tabular}
\caption{Common anatomy terms for filtering the search scope of domain-level context retrieval. This list is curated from the anatomy task ontology (Table~\ref{tab:anatomy_ontology}) and frequent section headers. Stop words are removed. }
\label{tab:common_anatomy}
\end{table*}

To filter for anatomy-informative sentences, we employ the same BM25 model to match the entire search corpus with a single anatomy string, which was cheaply curated from the anatomy normalization categories and frequently auto-extracted section headers, as shown in Table~\ref{tab:common_anatomy}. After filtering, the search corpus is reduced to 36\%, shrinking from 51,481 sentences to 18,959 sentences.

\section{Implementation details}
\label{sec:appendix_implementation}

The mSpERT models are trained at a batch size of 15 for 15 epochs.\footnote{We use full event schema for mSpERT models, including all attribute types in the annotations, including anatomy, characteristic, size, size-trend, and count.  While T5 models only extract the most important attribute, the anatomy attribute. } T5 models utilize a maximum input length of 768 tokens and a maximum output length of 512 tokens. When incorporating all types of contexts, we double the input maximum length to 1536 tokens. We train 15 epochs, with a batch size of 8. For the T5 large model, to accommodate a single NVIDIA A100 device, we employ gradient accumulation by using a batch size of 2 and accumulating four times.




\section{Case study for context benefits}
\label{sec:appendix_case_study}
\begin{table*}
    \centering
    \caption{Error examples with helpful contexts}
    \label{tab:case_study}
\begin{tabularx}{\textwidth}{|>{\hsize=.6\hsize}X|>{\hsize=1.0\hsize}X|>{\hsize=1.4\hsize}X|}
   \hline

        Error with example & Contexts & Before and after \\
        \hline
        [ambiguity] \textbf{Right middle lobe} nodule (4, 81) measures 3 mm, previously 4 mm & [document-level section header] Scattered bilateral \textbf{pulmonary} nodules, as described below  & before: Hepato-Biliary | Liver\newline after: Respiratory | Lung \\
        \hline
        [hard vocabulary] There is \textbf{biapical} fibrosis & [domain-level BM25] There is biapical \textbf{pulmonary} fibrosis compatible with radiation therapy & before: Musculo-Skeletal | Bone and or Joint  after: Respiratory Lung \\
        \hline
    \end{tabularx}
\end{table*}

We observe that contexts can aid in disambiguation (e.g. right middle lob) and understanding difficult medical terminology (e.g. biapical). For both examples presented in Table~\ref{tab:case_study},  contexts include the term "pulmonary", indicating the anatomies are related to lungs.

\end{document}